\documentclass{llncs}
\usepackage{amsmath,amssymb,amsfonts}
\usepackage{mathtools}
\usepackage{algorithmic}
\usepackage{graphicx}
\usepackage{textcomp}
\usepackage{xcolor}
\usepackage{todo}
\usepackage{mathtools}

\usepackage{color, colortbl}
\definecolor{LightGray}{gray}{0.95}
\definecolor{DarkGray}{gray}{0.5}
\usepackage{enumerate}
\usepackage{footnote}
\usepackage{subcaption}
\usepackage{array}
\usepackage{hhline}
\usepackage{float}
\usepackage{makeidx}  
\begin{document}
\frontmatter          
\pagestyle{headings}  
\addtocmark{Hardware Acceleration of an XNOR Traffic Signs Classifier} 
\mainmatter              
\title{Exploration of Hardware Acceleration Methods for an XNOR Traffic Signs Classifier}
\titlerunning{Hardware Acceleration of an XNOR Traffic Signs Classifier}  
%
\author{Dominika Przewlocka-Rus \and Marcin Kowalczyk \and Tomasz Kryjak}
%
\authorrunning{Dominika Przewlocka-Rus et al.} 
%
\tocauthor{Dominika Przewlocka-Rus, Marcin Kowalczyk and Tomasz Kryjak}
%
\institute{AGH University of Science and Technology in Krakow, Poland,\\
\email{\{dominika.przewlocka, marcin.kowalczyk, tomasz.kryjak\}@agh.edu.pl}}

\maketitle              

\begin{abstract}

Deep learning algorithms are a~key component of many state-of-the-art vision systems, especially as Convolutional Neural Networks (CNN) outperform most solutions in the sense of accuracy. 
To apply such algorithms in real-time applications, one has to address the challenges of memory and computational complexity. 
To deal with the first issue, we use networks with reduced precision, specifically a binary neural network (also known as XNOR). 
To satisfy the computational requirements, we propose to use highly parallel and low-power FPGA devices.
In this work, we explore the possibility of accelerating XNOR networks for traffic sign classification. 
The trained binary networks are implemented on the ZCU 104 development board, equipped with a Zynq UltraScale+ MPSoC device using two different approaches. 
Firstly, we propose a custom HDL accelerator for XNOR networks, which enables the inference with almost 450 fps. 
Even better results are obtained with the second method -- the Xilinx FINN accelerator -- enabling to process input images with around 550 frame rate.
Both approaches provide over $96\%$ accuracy on the test set.

\end{abstract}

\section{Introduction}
\label{sec:introduction}

The problem of traffic sign detection and recognition is important for many applications, such as advanced driver assistance and safety systems, autonomous vehicles (self-driving cars), or traffic signs maintenance (i.e., detection of damage or incorrect positions). 
Due to the wide application and specificity of the problem, three aspects should be considered:
(1) The algorithm has a~direct impact on the safety of both its user (e.g. driver) and other road users (e.g. pedestrians). 
Therefore, we propose an approach based on machine learning. 
In particular, we use deep convolutional neural networks, which are very well suited for solving classification problems and characterised by high accuracy;
(2) We expect the system to process images in real-time, since this has a direct impact on the so-called 'reaction time' and therefore also safety.
(3) We would like to keep the power consumption as low as possible, because the energy budget of modern cars (especially electric) is very limited.
Therefore, it is justified to use programmable heterogeneous systems that enable designing almost any computational architecture, also one that will process data in parallel and pipelined, with minimal energy overhead.

Since the hardware implementation of such computationally complex solutions is not straightforward, for the presented work, we (1) decided to use XNOR networks; (2) explore the possibilities of hardware acceleration with custom implementation as well as using open-source library from Xilinx -- FINN \cite{blott2018finn}, \cite{finn}.

The main contributions of this paper are: (1) the proposed energy efficient and real-time accurate traffic sign classifier run on the ZCU 104 development board with a heterogeneous Zynq UltraScale+ MPSoC device with clock frequency of 100 MHz; (2) design space exploration and comparison of different acceleration techniques (custom vs FINN).

The remainder of this paper is organised as follows: in Section \ref{sec:previous} we briefly summarise the previous work concerning the research topic. Section \ref{sec:training} describes the proposed network architectures as well as classification results. In Section \ref{sec:hw} we present the hardware accelerators' experiments -- in \ref{ssec:hw-custom} for the custom one, in \ref{ssec:hw-FINN} for the FINN-based. Section \ref{sec:results} discusses the obtained results and shows a~comparative analysis of both solutions. Finally, in Section \ref{sec:sum} we summarise our work and discuss possible future research directions.




\section{Previous work}
\label{sec:previous}

Traffic sign recognition or classification is a particularly interesting topic due to its application in autonomous vehicles (AV) or advanced driver assistant systems (ADAS).
One of the most popular databases is GTSRB (German Traffic Sign Recognition Benchmark), which was first presented for the challenge during the IJCNN 2011 conference \cite{ijcnn2011}.
The winners with the highest classification accuracy on the test set were: (1) 99.46\% \cite{ciresan2012}, (2) $98.31\%$ \cite{sermanet2011} and (3) $96.14\%$ \cite{zaklouta2011}.
The first two were based on convolutional neural networks, while the third used a random forests classifier. 
The human performance in that task was estimated as $98.84\%$.
Due to the popularity of the topic, several other papers on classification of traffic signs were presented later. 
The state-of-the-art classification accuracy using convolutional neural networks is about $99.71\%$ \cite{garcia2018} for GTSRB. 

The described in the literature hardware implementations of traffic sign recognition mainly focus on solutions without the use of deep convolutional neural networks.
In \cite{han2014} and \cite{farhat2016} the authors propose a~software-hardware implementation of traffic sign detection based on colour information in the HSV space and pattern matching based classification.
Article \cite{zhou2015} describes a~software-hardware solution based on the HOG (Histogram of Oriented Gradients) algorithm. 
In the work \cite{yao2017} a~hardware~implementation of a~deep convolutional neural network that classifies signs was proposed.
The implementation of binary neural networks in FPGAs is the subject of several articles. 
In \cite{Zhao2017} the authors summarise the first HLS binary network accelerator targeting architecture for CIFAR-10 dataset classification. 
Comparing to other hardware implementations (for standard convolutional networks), the binary one was far more efficient in terms of resources utilisation and power consumption, achieving even 6 times more GOPS/Watt.
In \cite{blott2018finn}, \cite{finn} the authors present the FINN tool for fast implementation of quantised neural networks and test their solution on common benchmarks, such as MNIST, CIFAR-10 or SVHN. 
\cite{Zhou2017} describes a solution for CIFAR-10 benchmark that achieves $332$ classifications per second with accuracy of $86.06\%$. 
Authors of \cite{Shimoda2017} propose a decomposition of the input maps (multichannel image) to $1$-bit precision, which boosts the calculations of the first convolutional layer. 
\cite{Liang2018} presents an implementation based on Resource-Aware Model Analysis and evaluates the accelerator on MNIST, CIFAR-10 and AlexNet benchmarks. 




\section{Traffic sign classification with XNOR networks}
\label{sec:training}

\begin{figure}[!t]
\centering
    \includegraphics[width=12cm]{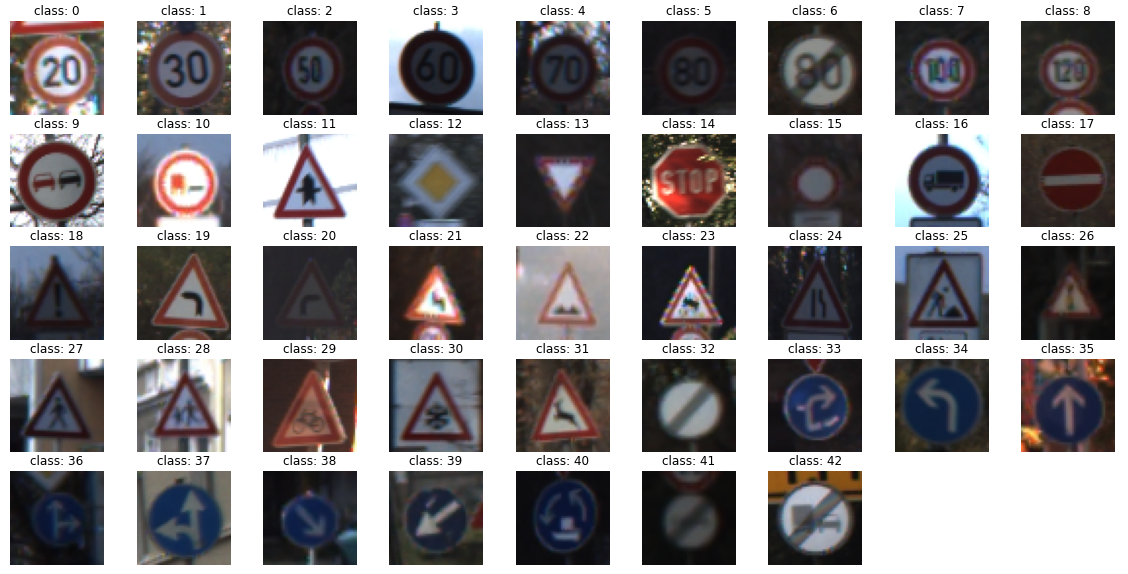}
    \caption{Sample raw data from the GTSRB dataset representing 43 classes, resized to square images for better readability. Note the high variability, especially lightning}%
    \label{fig:gtsrb-raw}%
\end{figure}

For the classification of traffic signs, the GTSRB (German Traffic Sign Recognition Benchmark) database was used \cite{ijcnn2011}.
It consists of images of German traffic signs representing 43~classes (Fig. \ref{fig:gtsrb-raw} -- for proper names of classes please refer to \cite{ijcnn2011}).
12630 images are used as the test set and 39209 as the training set, however, the number of training samples belonging to each class is not balanced.
Each image is in the RGB colour space of size varying between $15 \times 15$ up to $250 \times 250$ pixels.
At the same time, not all are square, differ in illumination and quality.
Due to the above described issues, the images had to be standardised before training. 
The preprocessing of the input data was split into two parts: (1) standardisation of images with CLAHE (Contrast Limited Adaptive Histogram Equalisation - \cite{zuiderveld1994}) for the Y channel from the YCbCr colour space and (2) data augmentation to balance class representations.
Data augmentation was performed using rotation, translation, shear, and scaling operations, so finally each class is represented by 10000 examples.

Candidates for traffic signs are classified by an~XNOR-type neural network, described in detail in the work \cite{Courbariaux16}. 
The network is characterised by binary weights and activations: $ W \in \{- 1, 1 \} \land f (Wx) \in \{- 1, 1 \} $ (with $W$ as weights, $x$ as layer's input and $f()$ as activation function). 
The input image is represented by integer or floating point values, which means that the first convolutional layer is not fully binary. 

The binary operations can vastly reduce the computational complexity of the network, which can be crucial in the case of embedded systems and particularly for the low-power ones. 
For a~binary convolutional layer, the output weighted sum can be calculated in three stages: 
\begin{enumerate}
\item XNOR operation between the vectorised input window $I_{in}$ and the vectorised filter $W$ -- $I_{XNOR} = I_{in} \bar{\oplus} W$. 
\item Bit count the obtained vector -- $P = count(I_{XNOR})$. 
\item Calculation of the convolution result based on $C = 2 * P - N$, where $ N $ is the size of the window.
\end{enumerate}
Next, the real-valued bias is added.
The activation is a binarisation function: $Pix_{out} = max{(0, min{(1, \frac{Pix_{in} + 1}{2}})}$.

Subsections \ref{ssec:custom-net} and \ref{ssec:FINN-net} describe the proposed networks accelerated using the ZCU 104 development board -- respectively, for the custom accelerator and the Xilinx FINN based one.

\subsection{Custom architecture}
\label{ssec:custom-net}

The proposed network architecture is presented in Table \ref {tab:netArch}. 
The convolutional block consists of the following layers (1) convolutional, (2) maxpooling, (3) batch normalisation, (4) activation. 
The dense (fully connected) block consists of the following layers (1) fully connected, (2) batch normalisation, (3) activation. 
The last dense block does not have an activation layer -- the vector obtained from the batch normalisation layer is the output vector (network classification result).

For training the network, we used the code provided by \cite{Courbariaux16}. 
The input images' pixels are normalised to range $[-1, 1]$ and the Square Hinge loss is used.
The network after $ 570 $ epochs achieved an accuracy of $96.3\%$ on the test set ($98.5\%$ on validation set).
\begin{center}
	\begin{table}[!t]
		\centering
		\caption {Neural network's architecture for custom accelerator}
		\begin{tabular}{ l  l  l  l }
		\hline\noalign{\smallskip}
			{Layer} & {Input size}  & {Output size} & {Kernel size}  \\
			\noalign{\smallskip}
			\hline
			\noalign{\smallskip}
			Conv-1 & $32 \times 32 \times 3$ &  $28 \times 28 \times 64$ & $5 \times 5 \times 64$ \\
			Max-1      & $28 \times 28 \times 64$ &  $14 \times 14 \times 64$ & $2 \times 2$ \\
			Conv-2    & $14 \times 14 \times 64$ &  $10 \times 10 \times 128$ & $5 \times 5 \times 128$ \\
			Max-2     & $10 \times 10 \times 128$ &  $5 \times 5 \times 128$ & $2 \times 2$ \\
			FC-1     & $5 \times 5 \times 128$ &  $512$ & $1 \times 1 \times 512$ \\
			FC-2     & $512$ &  $43$ & $1 \times 1 \times 43$ \\
			\hline
		\end{tabular}
		\label{tab:netArch}
	\end{table}
\end{center}

\subsection{FINN architecture}
\label{ssec:FINN-net}
Since the FINN tool is still under strong development, several features are not yet available. 
The architecture proposed in Section \ref{ssec:custom-net} cannot be transformed into hardware modules using FINN (for today).
It is required that the consecutive layers are designed in such way, that the output of the last convolutional one has a size of $1x1$. 
In table \ref{tab:netArch-FINN} the adjusted architecture is shown.

The input images are normalised to the range $[0, 1]$ and the accuracy for the test set of $96.24\%$ was achieved.
\begin{center}
	\begin{table}[!t]
		\centering
		\caption {Neural network's architecture for FINN tool}
		\begin{tabular}{ l  l  l  l }
		\hline\noalign{\smallskip}
			{Layer} & {Input size}  & {Output size} & {Kernel size}  \\
			\noalign{\smallskip}
			\hline
			\noalign{\smallskip}
			Conv-1 & $32 \times 32 \times 3$ &  $28 \times 28 \times 64$ & $5 \times 5 \times 64$ \\
			Max-1      & $28 \times 28 \times 64$ &  $14 \times 14 \times 64$ & $2 \times 2$ \\
			Conv-2    & $14 \times 14 \times 64$ &  $10 \times 10 \times 128$ & $5 \times 5 \times 128$ \\
			Max-2     & $10 \times 10 \times 128$ &  $5 \times 5 \times 128$ & $2 \times 2$ \\
			Conv-3     & $5 \times 5 \times 128$ &  $5 \times 5 \times 513$ & $1 \times 1 \times 512$ \\
			FC-1     & $512$ &  $43$ & $1 \times 1 \times 43$ \\
			\hline
		\end{tabular}
		\label{tab:netArch-FINN}
	\end{table}
\end{center}

\section{Hardware acceleration}
\label{sec:hw}

The hardware acceleration targets a~heterogeneous Zynq UltraScale+ MPSoC device, available on the ZCU 104 development board, both for custom and FINN accelerator. 
The target frequency is set to 100MHz (this value can be increased if necessary).
Subsection \ref{ssec:hw-custom} describes the developed accelerator as well as the obtained results for the proposed traffic sign classification network.
Subsection \ref{ssec:hw-FINN} summarises the experiments with the FINN tool.

\subsection{Custom accelerator}
\label{ssec:hw-custom}
\begin{figure}[h]
\centering
    \includegraphics[width=12cm]{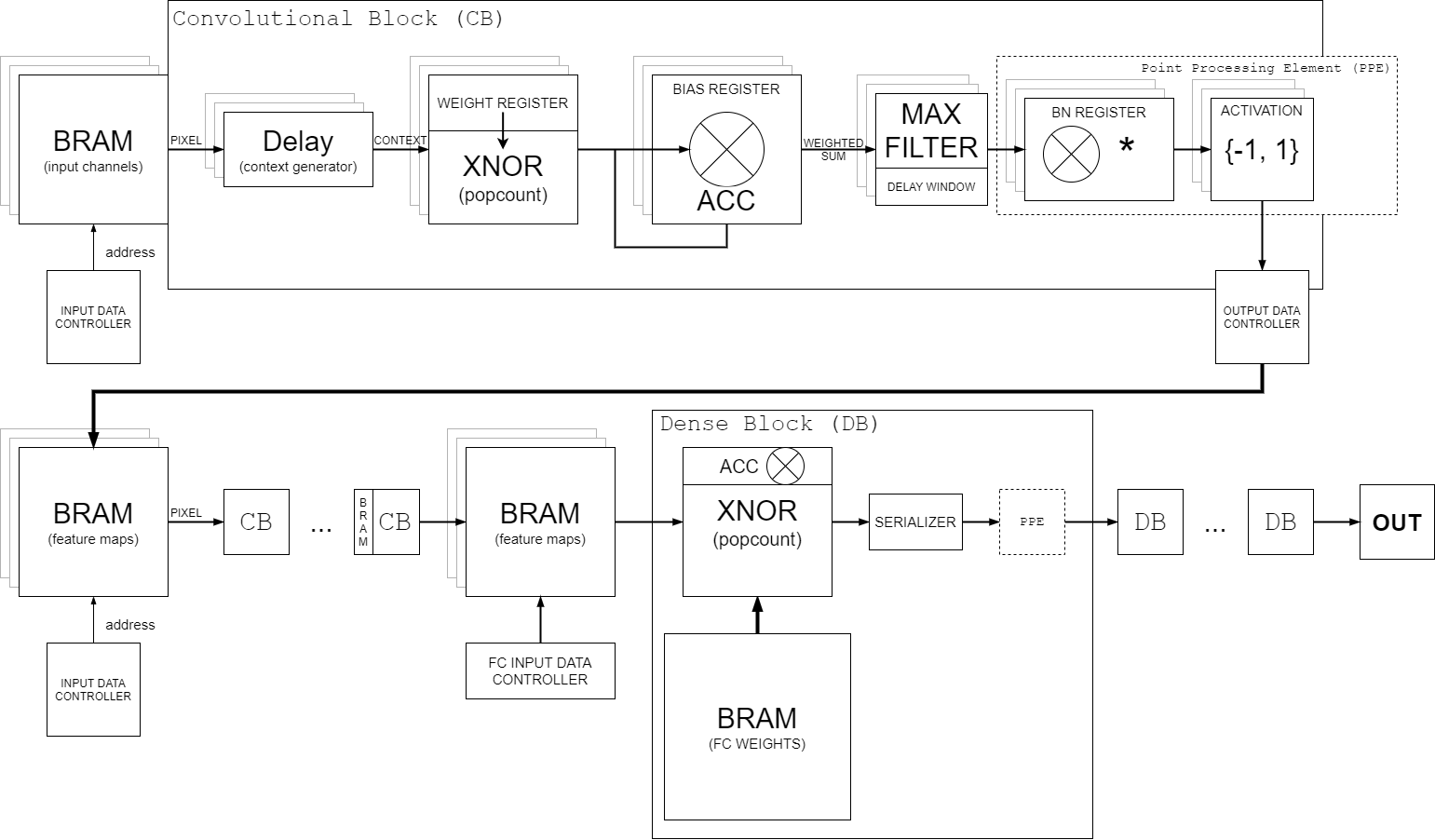}
    \caption{A simplified scheme of the custom accelerator. The proposed architecture consists of three main components: Convolutional Blocks, Dense Blocks and BRAM sets for inputs and outputs from layers. The operations in layers are streamlined.}%
    \label{fig:accarch}%
\end{figure}

The simplified accelerator scheme is presented in Fig. \ref{fig:accarch}. 
The proposed solution can be used for the acceleration of virtually any simple convolutional network (within the available resources limitation) using the developed Verilog code generator.

The key components of the proposed accelerator are the Convolutional Block and Dense Block parts, as well as several BRAM (Block RAM) sets which store the input, intermediate results and fully connected layers' weights.
The overall organisation of operations is based on several crucial ideas:
\begin{enumerate}
    \item The input and intermediate feature maps are organised in BRAM sets in such way that each BRAM stores one channel, which ease the process of writing the output feature maps.
    \item The data is read in a~sequence that enables the calculation of convolutional and maxpooling layers without additional memory usage.
    \item The weights of the convolutional layers are stored in the on-board registers for better access.
    \item The weights of the fully convolutional layers are stored in BRAM. For enhanced parallelisation of the calculations, the memory is organised in a way to enable reading at once the weights connected with the one, selected input.
\end{enumerate}
The Convolutional Block (CB) consists of modules responsible for layers: convolutional, maxpooling, batch normalisation, and binary activation.
In the case of the first convolutional layer, the network's input is a~greyscale (one channel -- this is applicable for the described TSR network) or colour image (three channels) -- each channel is stored in a~separate BRAM memory (each 16Kb). 
In the case of subsequent layers, the BRAM set stores the output (feature maps) from the previous layer. 
The filter weights are stored in the on-board registers, therefore the utilisation of BRAM memory is reduced, and the time needed to read the weights is shortened.
Reading pixels, from both the input image and the feature maps between successive convolutional blocks, is organised in a~specific order that allows the implementation of the maxpooling operation without the use of additional memory. 
The data is read as follows:
\begin{enumerate}
	\item Read one context along all channels, in relation to the same central pixel.
	\item Move the filter window and repeat the previous step -- further contexts are selected in such way, that in the result of successive convolutions, a~context ready for the maxpooling operation is obtained.
\end{enumerate}
All filters work in parallel, receiving the same input data -- contexts in a~specific position (with a~specific central pixel) along all input channels. The partial convolutions are accumulated and the weighted sum along the channels is calculated. 

The convolutional filter module varies with the input type. 
For the first layer -- real-valued input -- it is an accumulator that adds or subtracts incoming pixels to/from the weighted sum, depending on the weight value ($ -1 $ or $ 1 $). 
For a binary input, the convolution is performed in three stages:
\begin{enumerate}
	\item Vector of weights ('flattened' filter) is XNORed with the context (also in the form of a vector) -- the operation takes $ 1 $ clock cycle.
	\item The number of ones in the received vector is counted. Bit counting operation is performed using Hamming weights ($ 5 $ clock cycles).
	\item The result of the convolution $ C $ is calculated based on $ C = 2 * P - N $ ($ 1 $ clock cycle).
\end{enumerate}

The obtained partial results (from subsequent channels) are accumulated and the bias is added. 
The maxpooling module accepts consecutive pixels and when ready acts as maximum filter.
The output value is finally subjected to batch normalisation and binary activation (standard arithmetic operations). 
The batch normalisation formula is simplified to $Pix_{out} = A \cdot Pix_{in} + B$, where $A = \gamma \cdot \frac{1}{\sigma}$ and $B = - \gamma \cdot \frac{1}{\sigma} \cdot \mu + \beta$, in order to reduce the resource utilisation (DSP instances).
The results are written in parallel from all filters, to different, appropriate BRAM instances (corresponding output map).

Implementation of a fully connected layer (Dense Block (DB)) assumes the use of a~larger BRAM memory due to the significantly higher number of weights (in relation to the convolutional layer).
The input pixels are read in a~serial way from the feature map of the previous convolutional layer. 
Before receiving a~pixel, a~set of weights corresponding to the connection for all neurons is read from the BRAM memory -- one word in memory stores the weights of a~given connection for all the layer's neurons. 
The input is XNORed with each weight (i.e., the equivalent of multiplication in networks with full precision is performed) and then the results are sent to the respective accumulators representing the neurons. 
The weighted sum of all the layer's neurons is calculated in parallel. 
Then the data (outputs from all neurons) is serialised, bias is added, and values are normalised, if needed.

The post-implementation resource usage is presented in Tab. \ref{tab:netResources}. 
The frame rate of network's inference is around $449.25$ (calculated basing on clock frequency and number of clock cycles needed for input processing, as well as tested in hardware), with power consumption of $4.396$W (estimated using Vivado). 
\begin{center}
	\begin{table}[!t]
		\centering
		\caption{Post-implementation resources usage for Binary Convolutional Neural Network on ZCU104}
		\begin{tabular}{ l  l  l  l }
		\hline\noalign{\smallskip}
			{Resource} & {Utilisation}  & {Available} & {Utilisation \%}  \\
			\noalign{\smallskip}
			\hline
			\noalign{\smallskip}
			LUT & $79302$ &  $230400$ & $34.42$ \\
			LUTRAM & $350$ &  $101760$ & $0.34$ \\
			FF & $44170$ &  $460800$ & $9.59$ \\
			BRAM & $152$ &  $312$ & $48.72$ \\
			DSP & $902$ &  $1728$ & $52.20$ \\
			\hline
		\end{tabular}
		\label{tab:netResources}
	\end{table}
\end{center}

\subsection{FINN accelerator}
\label{ssec:hw-FINN}

Acceleration of neural networks using the FINN tool requires several well-defined steps. 
Given the model, preferably trained using the Brevitas \cite{brevitas} tool, in the ONNX (Open Neural Network Exchange) format, one has to perform a set of operations from streamlining transformations, converting to HLS (High-Level Synthesis) layers and dataflow partitioning, to adjusting folding to maximise the performance. 
All but the last is done in a rather automatic manner and for more information we suggest consulting the FINN documentation. 
The folding influences the frame rate of the network and the usage of resources. 
Finally, the HLS IP per layer can be generated, as well as the stitched design and Vivado/Vitis project ready to be synthesised and run in hardware. 
If the accelerated network matches the framework's capabilities, the process is a rather smooth one.

To explore the possibilities of FINN, we decided to perform several experiments with folding, following the official examples -- 'there are several folding factors for each layer, controlled by the PE (parallelisation over outputs) and SIMD (parallelisation over inputs) parameters (...) the higher the PE and SIMD values are set, the faster the generated accelerator will run, and the more FPGA resources it will consume'.
The experiments' parameters, as well as the final frame rate of the accelerated network and energy consumption are shown in Table \ref{tab:finn-folding} -- for each layer, the folding parameters (PE, SIMD, FIFO depth) are presented.
Table \ref{tab:finn-resources} shows the resources' usage for the defined experiments. 
It is easily observed that with higher folding parameters, the network  process the input faster -- the difference between experiments 1 and 4 is more than 500 frames per second. 
One can also notice the rather slight increase of the used resources -- perhaps for more complex (in terms of architecture) networks the results would vary more.
\begin{center}
	\begin{table}[!t]
		\centering
		\caption{Folding experiments for the TSR network with the FINN accelerator -- parameters, final throughput and energy consumption (estimated with Vivado).}
		\begin{tabular}{  l  c  c  c  c  c  c  }
		\hline\noalign{\smallskip}
			No. & Conv 1 & Conv 2 & Conv 3 & FC 1 & FPS & Power [W] \\
			\noalign{\smallskip}
			\hline
			\noalign{\smallskip}
			$1$ & $(4, 1, 32)$ & $(4, 4, 32)$ &  $(4, 4, 32)$ & $(1, 1, 32)$ & $64.51$ & $3.521$ \\
			$2$ & $(4, 1, 128)$ & $(4, 4, 128)$ &  $(4, 4, 128)$ & $(1, 1, 128)$ & $64.51$ & $3.523$ \\
			$3$ & $(8, 1, 128)$ & $(8, 8, 128)$ &  $(8, 8, 128)$ & $(1, 1, 128)$ & $212.41$ & $3.522$ \\
			$4$ & $(16, 1, 128)$ & $(16, 16, 128)$ &  $(4, 16, 128)$ & $(1, 1, 128)$ & $582.22$ & $3.547$ \\
			\hline
		\end{tabular}
		\label{tab:finn-folding}
	\end{table}
\end{center}
\begin{center}
	\begin{table}[!t]
		\centering
		\caption{Post-implementation resources usage for the FINN accelerated TSR network in ZCU104}
		\begin{tabular}{ l  c  c  c  c  c}
		\hline\noalign{\smallskip}
			{Resource} & { Total Available} & {E1 [\%]} & {E2 [\%]} & {E3 [\%]} & {E4 [\%]} \\
			\noalign{\smallskip}
			\hline
			\noalign{\smallskip}
			LUT & $230400$ & $5.81$ & $5.89$ & $5.75$ & $6.14$ \\
			LUTRAM & $101760$ & $1.62$ & $1.67$ & $1.59$ & $1.70$ \\
			FF & $460800$ & $4.29$ & $4.29$ & $4.4$ & $4.62$ \\
			BRAM & $312$ & $22.92$ & $22.92$ & $22.28$ & $21.96$ \\
			DSP & $1728$ & $-$ & $-$ & $-$ & $-$ \\
			\hline
		\end{tabular}
		\label{tab:finn-resources}
	\end{table}
\end{center}

\section{Comparison of the considered approaches}
\label{sec:results}

Firstly, it should be noted that when starting the work on the custom XNOR accelerator, we were not familiar with the FINN tool, as it was in an early stage of development back then.
After completing our implementation, we have read about FINN and only then decided to compare both approaches.
The analysis of the results shall be divided into two parts -- the binary networks and their accuracy, and the hardware acceleration.


In Section \ref{sec:training} we have proposed two network architectures with reduced precision for traffic sign recognition on the GTSRB dataset -- the 'original' and FINN-adapted. 
Both networks achieved similar and relatively good accuracy, despite the use of different tools.
We have compared our networks to other traffic sign classifiers and placed ourselves among the existing top CNN-based classifiers -- Tab. \ref{tab:state-of-the-art}.
In \cite{garcia2018} and \cite{ciresan2012} the authors proposed networks with significantly more parameters. 
In \cite{sermanet2011} the presented solution employs layer skipping to provide different scales of receptive fields to the classifier, which also increases the model complexity. 
All mentioned solutions work with floating point precision -- when comparing the accuracy of our XNOR models with other proposals, the 'size issue' and hardware acceleration suitability should be also taken into account. 
\begin{center}
	\begin{table}[!t]
		\centering
		\caption{Comparison of our traffic sign classifier to the state-of-the-art solutions. C stands for the network developed with the custom accelerator, F -- for FINN}
		\begin{tabular}{lcc}
		\hline\noalign{\smallskip}
			Network & Accuracy &  \\
			\noalign{\smallskip}
			\hline
			\noalign{\smallskip}
			Garcia et al. \cite{garcia2018} & $99.71\%$ & \\
            Ciresan et al. \cite{ciresan2012} & $99.46\%$ & \small{1st at IJCNN 2011} \\
            Sermanet et al. \cite{sermanet2011} & $99.17\%$ & \small{2nd at IJCNN 2011} \\
            Human performance & $98.84\%$ & \\
            \textbf{Our XNOR Net -- C} & $96.3\%$ & \\
            \textbf{Our XNOR Net -- F} & $96.24\%$ & \\
            Zaklouta et al. \cite{zaklouta2011} & $96.14\%$ & \small{3rd at IJCNN 2011} \\
			\hline
		\end{tabular}
    \label{tab:state-of-the-art}
	\end{table}
\end{center}

The results presented in Section \ref{sec:hw} basically prove the lead of the FINN accelerator, both in terms of energy consumption and the used resources (Tab. \ref{tab:finn-folding} and \ref{tab:finn-resources}) for the same test conditions -- the ZCU104 device and clock frequency of 100 MHz.
The proposed custom accelerator needs over $4.396W$, which is over $25\%$ of the total power consumption of the FINN-generated one. 
The disproportion in resource usage is also noticeable.
The inference speed of the FINN-accelerated network (Tab. \ref{tab:finn-folding}) depends on the folding parameters. 
However, achieving the frame rate close to the one from the custom accelerator neither is a problem nor does it cause a significant increase in power demand -- which is still lower for the FINN-based accelerator.

Perhaps it is worth noting what is the cause of such results -- experiment E4 (see Tab \ref{tab:finn-folding} and \ref{tab:finn-resources}) is chosen as a reference point. 
Since the resources' usage and energy consumption are closely related, a comparative analysis of the architectures of these two accelerators should help to understand the differences.
Firstly, one can observe a great difference in the BRAM usage ($48.72\%$ vs $21.96\%$). 
The main cause shall be storing the outputs of the subsequent convolutional blocks in case of the custom accelerator. 
The FINN architecture is fully streamlined and consecutive outputs are immediately passed for processing to the next layer (without any buffering).
Next, one can notice the lack of usage of DSP elements in FINN accelerator. 
For the custom one, DSP are used as neurons in the fully connected layers -- in case of the FINN tool it seems that the accumulation is implemented with popcount operation.
The last factor influencing the number of used LUTs ($34.42\%$ vs $6.14\%$) and FFs ($9.59\%$ vs $4.62\%$) could be in the batch normalisation module, which is implemented as threshold in the FINN accelerator -- in contrary to the arithmetic operations in the custom one.

\section{Summary}
\label{sec:sum}

The starting point for this work was the development of an energy-efficient and real-time traffic sign classification system. 
The proposed neural networks are characterised by high accuracy and at the same time relatively low computational complexity -- both in terms of the architectures' size and the ability to be inferenced with effective binary operations.
As a consequence, the hardware acceleration on the ZCU 104 developement board with Zynq UltraScale+ MPSoC device was possible.

The exploration of acceleration methods was focused on two approaches: using the designed custom accelerator and the FINN tool from Xilinx.
The conducted experiments showed the FINN tool superiority, due to the use of some advanced optimisations, which we did not apply in our custom solution.
However, it should be once more emphasised that our research goal was not to propose a better solution than FINN, as we did not know of its existence or any other open-source XNOR accelerator (at the beginning).

Eventually, using the FINN accelerator, we achieved the set goal, which was an energy-efficient real-time classifier. 
The developed classifier is ready to use in the complete embedded traffic sign recognition system.
Because it is characterised by high frame rate, it is also suitable for UHD input video stream.
The development of the system for both recognition and classification for high resolution camera input (up to 4K) shall be continued in future work.

\section*{Acknowledgements}
The work presented in this paper was supported by the National Science Centre project no. 2016/23/D/ST6/01389 entitled ``The development of computing resources organisation in latest generation of heterogeneous reconfigurable devices enabling real-time processing of UHD/4K video stream''.

\end{document}